# Efficient Kernel-based Subsequence Search for User Identification from Walking Activity

Candelieri A., Fedorov S., Messina E.

University of Milano-Bicocca, Department of Computer Science, Systems and Communication

**Abstract**

This paper presents an efficient approach for subsequence search in data streams. The problem consists in identifying coherent repetitions of a given reference time-series, eventually multi-variate, within a longer data stream. Dynamic Time Warping (DTW) is the metric most widely used to implement pattern query, but its computational complexity is a well-known issue. In this paper we present an approach aimed at learning a kernel able to approximate DTW to be used for efficiently analyse streaming data collected from wearable sensors, reducing the burden of computation. Contrary to kernel, DTW allows for comparing time series with different length. Thus, to use a kernel, a feature embedding is used to represent a time-series as a fixed length vector. Each vector component is the DTW between the given time-series and a set of "basis" series, usually randomly chosen. The vector size is the number of basis series used for the feature embedding. Searching for the portion of the data stream minimizing the DTW with the reference subsequence leads to a global optimization problem. The proposed approach has been validated on a benchmark dataset related to the identification of users depending on their walking activity. A comparison with a traditional DTW implementation is also provided.

*Keywords*: Subsequence Search on Streaming Data, Dynamic Time Warping, Kernel

## 1. Introduction

Dynamic Time Warping (Muller, 2007) is a technique to find the optimal alignment between two time-series, by considering the possibility to "warp" non-linearly one time-series by stretching or shrinking it along its time axis. The entity of the warping needed for the alignment is then used as a measure of the similarity/difference between the two time-series. A typical application of DTW is in speech recognition (Han et al., 2018), with the aim to determine if two waveforms represent the same spoken phrase. In a speech waveform, the duration of each spoken sound and the interval between sounds are permitted to vary, but the overall speech waveforms must be similar in terms of "shape". In addition to speech recognition, DTW has also been found useful in many other disciplines, including, gesture recognition (Tang et al., 2018), robotics (Calli et al., 2018), manufacturing (Bauters et al., 2018) and health-monitoring (Palyarom et al., 2009; Varatharajan et al., 2017; Candelieri et al., 2019). Measuring the similarity between two time-series is a core task for time-series clustering (Candelieri et al., 2014), where both data representation and pre-processing are critical choices, as well definition of a suitable similarity measure. Recently, in (Izakian et al., 2015) a fuzzy-clustering approach for time-series data has been proposed, where DTW was used as distance for comparing pairs of time-series.

Although its widely adoption in many application domains, a well-known issue of DTW is its computational complexity. Computing DTW between two time-series requires $O(NM)$, where $N$ and $M$ are the lengths of the two time-series. Thus, comparing a reference pattern to a large data stream (i.e., $M \gg N$), as well as searching for all the repetitions of the reference pattern within the stream, might be significantly computational expensive. Previous works addressing this issue are quoted in Section 2.

The specific contribution of this paper is:

- designing a kernel learning task aimed at approximating DTW to reduce computational burden of the pattern query task;
- reviewing and implementing a DTW based pattern query solution to compare effectiveness and efficiency of the proposed kernel-based method;
- validating the proposed approach on a benchmark dataset, namely the "User Identification From Walking Activity" dataset, freely downloadable from the UCI Repository:
  https://archive.ics.uci.edu/ml/datasets/User+Identification+From+Walking+Activity

The rest of the paper is organized as follows. Section 2 provides the methodological background about DTW, its recent innovations and applications, as well as computational drawbacks in the specific case of subsequence search. Section 3 describes how to learn a kernel to approximate DTW and, consequently, increase the computational efficiency in the search of repetitions of a specific pattern within long data streams. Section 4 presents the experimental setting and, more precisely, the benchmark dataset to validate the approach. Finally, Section 5 summarizes the results.

## 2. Background

### 2.1. Dynamic Time Warping

The core component of DTW is a data structure named *accumulated cost matrix*, denoted by $D \in \mathbb{R}^{N \times M}$ with $N$ and $M$ are the lengths of the two time-series to be compared, namely, $X = (x_1, \ldots, x_N)$ and $Y = (y_1, \ldots, y_M)$. Every entry of this matrix is computed as follows:

$$D_{i,j} = \min\{D_{i-1,j-1}, D_{i-1,j}, D_{i,j-1}\} + c(x_i, y_j) \quad \forall\, i = 1, \ldots, N \text{ and } j = 1, \ldots, M \tag{1}$$

where $c(\cdot,\cdot)$ is a *cost function* and $x_i$ and $y_j$ are the $i$th and $j$th values of $X$ and $Y$, respectively. The initialization can be simplified by extending the accumulated cost matrix $D$ with an additional row and column, specifically: $D_{i,0} = \infty$ and $D_{0,j} = \infty$. Then the recursion in the equation (1) holds for $i = 1, \ldots, N$ and $j = 1, \ldots, M$.

In the general case the two time-series could be multi-variate and, consequently, $x_i$ and $y_j$ could be vectors. The most widely adopted cost function is simply the Euclidean distance. Furthermore, the accumulated cost matrix satisfies, by construction, the following identities:

$$D_{i,1} = \sum_{k=1}^{i} c(x_k, y_1) \quad \forall\, i = 1, \ldots, N \tag{2}$$

and

$$D_{1,j} = \sum_{k=1}^{j} c(x_1, y_k) \quad \forall j = 1, \ldots, M \tag{3}$$

A *warping path* is defined as a sequence $p = \{p_1, \ldots, p_L\}$ of positions in $D$, where $p_l = (i_l, j_l)$, and defining an alignment between $X$ and $Y$ by assigning the element $x_{i_l}$ of the first series to the element $y_{j_l}$ of the second series. A warping path must satisfy the following conditions:

- Boundary condition: $p_1 = (1,1)$ and $p_L = (N, M)$. This condition enforces that alignment refers to the entire series, therefore the first elements between $X$ and $Y$, as well as the last, must be aligned to each other.
- Monotonicity condition: $i_1 \leq i_2 \leq \ldots \leq i_L$ and $j_1 \leq j_2 \leq \ldots \leq j_L$. This condition simply assures that if an element in $X$ precedes a second one this should also hold for the corresponding elements in $Y$, and vice versa.
- Step size condition: $p_{l+1} - p_l \in \{(1,0), (0,1), (1,1)\}$ for $l = 1, \ldots, L - 1$. This condition assures that no element in $X$ and $Y$ can be omitted and there are no replications in the alignment, meaning that all the index pairs contained in a warping path are pairwise distinct. Note that the step size condition implies the monotonicity condition.

The *total cost* associated to a warping path $p$ between $X$ and $Y$, and denoted by $c_p(X, Y)$, is computed as:

$$c_p(X, Y) = \sum_{l=1}^{L} c(x_{i_l}, x_{i_l})$$

An *optimal warping path* between $X$ and $Y$ is a warping path $p^*$ having minimal total cost over all the possible warping paths. Therefore, the DTW distance between $X$ and $Y$ is obtained as:

$$DTW(X, Y) = c_{p^*}(X, Y) = \min_{p}\{c_p(X, Y)\} \tag{4}$$

It is important to highlight that the DTW is symmetric if the cost function $c(\cdot,\cdot)$ is symmetric. However, DTW is in general not positive definite and generally does not satisfy the triangle inequality.

The following figure provides an example of accumulated cost matrix and the associated optimal warping path. The closer the optimal warping path to the diagonal, the lower the mis-alignment between the two series.

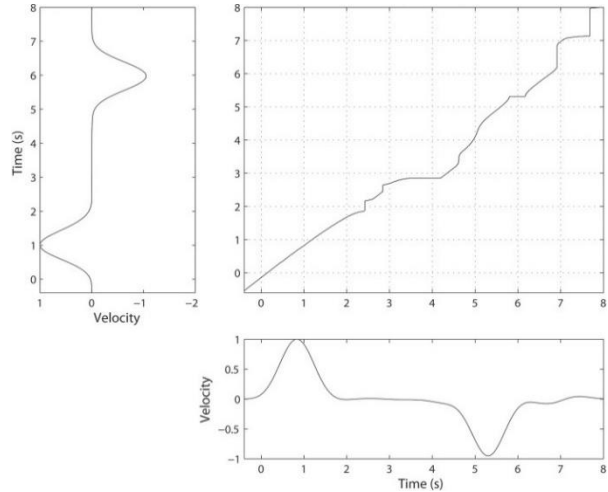

**Fig.1.** An illustration of accumulated cost matrix and associated optimal warping path when using DTW to align the two time-series in the picture.

The optimization problem (4) is solved by dynamic programming, with complexity $O(N, M)$. The programming algorithm is synthetized as follows:

**Optimal warping path algorithm**

**Input**: Accumulated cost matrix $D$
**Output**: optimal warping path $p^*$
1:  $i = N, j = M, p_1 = (i, j), l = 0$
2:  while( $i \neq 1$ or $j \neq 1$ ) {
3:      $l = l + 1$
4:      if ( $i == 1$ )
5:          $p_l = (i, j - 1)$
6:      else
7:          if( $j == 1$ )
8:              $p_l = (i - 1, j)$
9:          else
10:             $p_l = \text{argmin } \{D_{i-1,j-1}, D_{i-1,j}, D_{i,j-1}\}$
11: }
12: $p^* = \text{reverse}(p)$

The reverse operation, at the end of the algorithm, is necessary because the length $L$ of the optimal warping path $p^*$ is unknown a priori. Indeed, the optimal warping path is computed, according to the dynamic programming paradigm, in a reverse order starting from the position $(N, M)$ to the position $(1,1)$. Therefore, the reverse operation allows to give, as output, the optimal warping path as a sequence of positions coherent with the initial definition.

A commonly adopted DTW variant is to impose global constraint conditions on the admissible warping paths, with the aim to prevent undesired alignments by controlling the route of a warping path. Two widely adopted global constraints are the Sakoe-Chiba band (Sakoe et al., 1990) and the Itakura parallelogram (Itakura, 1975). Further to prevent undesired alignments, global constraints also allow to speed up DTW computation, because they basically limit the possible value of $L$, that is the length of the optimal warping path $p^*$.

The basic drawback of DTW is its computational cost, specifically when at least one of the two time-series is very long. Some preliminary studies addressed this issue through multi-level computation of DTW, as proposed in (Salvador & Chan, 2007). More in detail, the two time-series are initially sampled down to a very low resolution, a warping path is found at that resolution and then "projected" onto an incrementally higher resolution. This process of refining and projecting is continued until a warping path is identified for the full resolution time-series.

The brief review of the research efforts in optimizing both the efficiency and effectiveness of both the basic DTW algorithm, and of the higher-level algorithms that exploit DTW such as similarity search, clustering and classification is presented in (Mueen et al, 2016). It is discussed variants of DTW such as constrained DTW, multidimensional DTW and asynchronous DTW, and optimization techniques such as lower bounding, early abandoning, run-length encoding, bounded approximation and hardware optimization. Some relevant to desired task optimizations of DTW are presented in (Silva et al, 2016), an example of approximation accumulated cost matrix, and (Silva et al, 2018) speed up the distance calculations for univariate DTW. In (Keogh et al, 2001) can be found the exotic approach to increase robustness of similarity measure by constructing matrix over approximation of derivatives of neighbourhood samples, namely Derivative DTW. Finally, with respect to the topic of DTW approximation via kernel, (Nagendar and Jawahar, 2015) proposes a kernel aimed at learning the principal global alignments for the given data by using the hidden structure of the alignments from the training data. This approach is presented as more computationally efficient when compared to previous kernels on DTW distance, such as GA kernel (Cuturi et al., 2007; Cuturi, 2011) and Gaussian DTW kernel (Bahlmann et al., 2002).

### 2.2. Dynamic Time Warping for subsequence search

The problem of *pattern query*, based on DTW, is described in (Müller, 2007), and is also known as *subsequence DTW*. In the pattern query task, the time-series to be compared are characterized by significant difference in their lengths, i.e. $M \gg N$. The shortest one is *reference pattern*, represents a specific sequence to be search for within the longest one. Indeed, instead of searching for a global alignment between the two series, the goal is to find - at least - one subsequence of the reference pattern within the longer sequence, with optimal fitting (i.e., minimal DTW).

As in the basic DTW, we start from two time-series, $X = (x_1, ..., x_N)$ and $Y = (y_1, ..., y_M)$, but in this case $M \gg N$, meaning that $X$ is the reference pattern while $Y$ is the data stream where a subsequence of $X$ is searched for.

Let us denote by $a^*$ and $b^*$ the indices representing the begin and the end of the subsequence within $Y$, with $1 \leq a^* \leq b^* \leq M$. These indices are identified by solving the following optimization problem:

$$(a^*, b^*) = \underset{(a,b):1 \leq a^* \leq b^* \leq M}{\mathrm{argmin}} DTW(X, Y_{a:b}) \qquad (5)$$

where $Y_{a:b}$ is the subsequence $(y_a, ..., y_b)$ in $Y$.

Optimization problem (5) can be solved by applying a modification in the initialization of the previous DTW algorithm, consisting in replacing (3) with:

$$D_{1,j} = c(x_1, y_j).$$

In other words, contrary to the identities provided in Eq. 3, the starting position of the subsequence $a^*$ does not provide any value, except its own cost, and given this fact the cost of positioning $b^*$ depends only from DTW between reference pattern and chosen subsequence. The remaining values of the accumulated cost matrix $D$ are defined as in the basic DTW algorithm.

The index $b^*$ is determined as $b^* = \mathrm{argmin}_{b=1,...,M} D_{N,b}$. In case $b^*$ is not unique, the lexicographic order can be used to select among the multiple choices. Given the value $b^*$, then $a^*$ is obtained by applying the optimal warping path algorithm, starting from the position $(N, b^*)$. Finally, the resulting optimal warping path $p^* = (p_1, ..., p_L)$ must be reduced to $(p_l, ..., p_L)$, where $p_l$ is the maximum index such that $p_l = (a^*, 1)$, with $l \in \{1, ..., L\}$. Therefore, the optimal warping path between $X$ and $Y_{a^*:b^*}$ is given by $(p_l, ..., p_L)$; roughly speaking, all the elements preceding $y_{a^*}$ and those following $y_{b^*}$ are not considered in the alignment and, consequently, does not account for additional costs to DTW.

In the following we summarize how the subsequence search algorithm can be extended to find multiple repetitions of the reference pattern $X$ within the longer data stream $Y$. First, we introduce, as reported in (Müller, 2007), the distance function $\Delta: [1:M] \to \mathbb{R}$, with $\Delta(b) = D_{N,b}$ $b = 1, ..., M$, which assigns the minimal DTW that can be computed between the reference pattern $X$ and a subsequence in $Y$ ending in $y_b$. Given $b$, the starting index $y_a$ of the searched subsequence, is identified through the optimal warping path algorithm revised for subsequence search.

The step 8 of the algorithm is particularly important. Since the basic elements of the accumulated cost matrix are based on the Euclidian distance, we set $\Delta(b) = \infty$ in the neighbourhood of the optimal value $b^*$ to avoid already found subsequence (optimization process in step 4) as well as pathological cases of very short time series in the neighbourhood of one.

| Subsequence search algorithm |
|---|
| **Input**: reference pattern $X = (x_1, \ldots, x_N)$, a longer data stream $Y = (y_1, \ldots, y_M)$, with $M \gg N$, and a threshold $\tau$ |
| **Output**: a list $\mathcal{L}$ of repetitions of $X$ within $Y$ having, individually, a DTW lower than $\tau$. The list is ranked depending on the individual DTW |
| 1:    $\mathcal{L} = \emptyset$ |
| 2:    compute the accumulated cost matrix $D$ between $X$ and $Y$ |
| 3:    compute the distance function $\Delta$ |
| 4:    find $b^* = \text{argmin}_{b \in \{1, \ldots, M\}} \Delta(b)$ |
| 5:    if $(\Delta(b^*) > \tau)$ then STOP |
| 6:    find $a^*$ by using **Optimal warping path algorithm** but initialized in $j = b$ instead of $j = M$ |
| 7:    updating $\mathcal{L}$ as: $\mathcal{L} = \mathcal{L} \cup Y_{a^*:b^*}$ |
| 8:    Set $\Delta(b) = \infty$ for every $b$ in a suitable neighbourhood of $b^*$ |
| 9:    GO TO STEP 3 |

With respect to subsequence search, (Sakurai et al., 2005) proposes a fast search method that guarantees no false dismissals in similarity query processing and efficiently prunes a significant number of the search candidates, leading to a reduction in the search cost.

## 3. Learning a kernel to approximate DTW

### 3.1. Time-Series Kernels via Alignments

A number of global alignment kernels have been proposed in literature with the aim to extend DTW to kernel-based estimation method. The underlying idea is to avoid the problem of searching exactly the optimal warping path but to learn a kernel approximating the DTW value between two time series. Kernel methods have great promise for learning complex models by implicitly transforming a simple representation, like mapping typical Euclidian distance into a high-dimension feature space (Yen et al., 2014). The main obstacles for applying usual kernel method to time series are due to two distinct characteristics of time series: *(a)* variable length and *(b)* dynamic time scaling and shifts. Furthermore, direct use of DTW leads to a not positive definite kernel that does not provide a convex optimization problem (Cuturi et al., 2007). To overcome these obstacles, a family of global alignment kernels have been proposed by taking soft-max over all possible alignments in DTW to give a positive definite kernel (Cuturi et al., 2007, Cuturi, 2011, Marteau and Gibet, 2015). However, the effectiveness of the global alignment kernels is impaired by the diagonal dominance of the resulting kernel matrix proportional to the difference in the size between studied time series (Cuturi, 2011), which is the case of subsequence search. In addition, the quadratic complexity in length of time series makes it hard to scale (Cuturi et al., 2007). In (Wu et al., 2018) a random features mapping method for time-series embedding is proposed: the idea is to use an explicit mapping to represents any time-series as alignments to a set of randomly chosen "basic" times-series, having small length to significantly reduce computational cost. Starting from similar considerations, our approach aims at learning a kernel, based on random features mapping, to use for efficiently solve the subsequence search problem.

### 3.2. Learning a kernel for subsequence search

We assume to have two time-series, $X = (x_1, \ldots, x_N)$ and $Y = (y_1, \ldots, y_M)$, where $X$ is the "reference pattern" while $Y$ is a longer data stream where a subsequence $X$ is searched for into, i.e. $M \gg N$. Now we can also extend it to multimodal data with dimension $d$, where we have $X \in \mathbb{R}^{d \times N}$ and $Y \in \mathbb{R}^{d \times M}$.

A set of $R$ "basis" time-series is randomly generated: $S = \{s_1, s_2, \ldots, s_R\}$, where $s_i \in \mathbb{R}^{d \times D_i} \, \forall \, i = 1, \ldots, R$ and $L_i \in [L_{min}, L_{max}]$ is the length of the $i$-th time-series $s_i$, usually $L_i \ll M$ and $L_i < N$, and where $L_{min}$ and $L_{max}$ are the minimum and maximum length allowed. $R$, $L_{min}$ and $L_{max}$ are parameters of the algorithm and must be tuned manually. According to (Wu et al., 2018), if the time-series set $S$ is generated with normal distribution it shows good performance in further construction of kernel. Let us to define the *feature map* $\boldsymbol{\phi}_S(X) = (\phi_{s_1}(X), \ldots, \phi_{s_R}(X))^T$, where the $i$-th component of $\boldsymbol{\phi}_S(X)$ is the alignment between original time series $X$ and the random series $s_i$. We consider DTW as measure of this alignment, thus we must compute accumulated cost matrix $D$, discussed above, for every entry of the feature vector:

$$\boldsymbol{\phi}_S(X) = \big(DTW(X, s_1), \ldots, DTW(X, s_R)\big)^T. \tag{6}$$

The mapping (6) provides an $R$ dimensional vector without correspondence to dimensionality of original time series, hence it is used in the further construction of kernel able to work with time series of different length.

Although DTW must be computed $R$ times, the computational cost is reduced due to the reduced length of each $s_i$. Indeed, the cost for computing $\phi_{s_i}(X)$, in the worst case, is $O(NRD_{max})$: computational cost can be kept reasonable if $RD_{max} < N$; $RD_{max} \ll M$. Moreover, parallelization can be used to further improve efficiency, since the computation of each component of the vector $\boldsymbol{\phi}_S(X)$ is "embarrassingly" parallel, as depicted in the following figure, where we assume to have $R$ different processors.

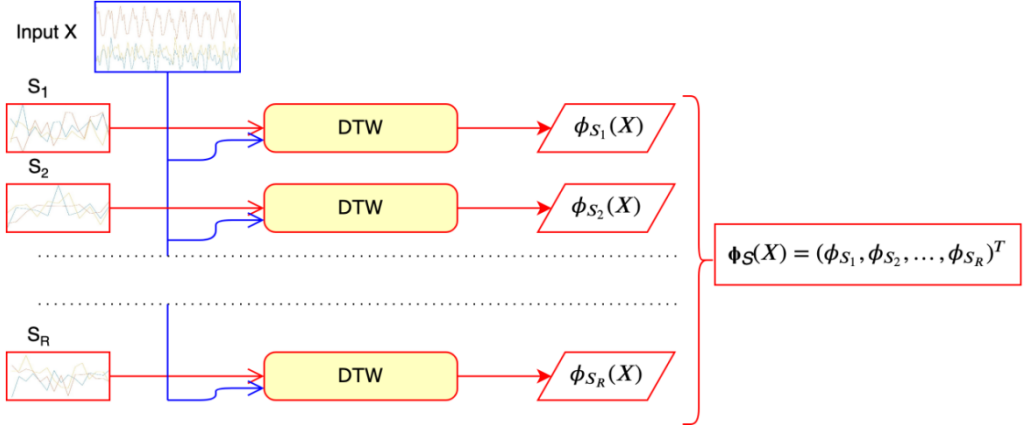

**Fig.2.** Illustration of the parallel computation of the components of the vector $\boldsymbol{\phi}_S(X)$. Time-series in the figure are assumed multi-variate

Given the two time-series $X$ and $Y$, $\boldsymbol{\phi}_S(X)$ and $\boldsymbol{\phi}_S(Y)$ are their associated representation in the space spanned by the their DTW with respect to the set of basis series $S$. Whichever is the length of $X$ and $Y$, their mappings $\boldsymbol{\phi}_S(X)$ and $\boldsymbol{\phi}_S(Y)$ have the same length, that is $R$. Contrary to (Wu et al., 2018) we decide to use a non-linear kernel – i.e., Radial Basis Function (RBF) kernel – to compare $\boldsymbol{\phi}_S(X)$ and $\boldsymbol{\phi}_S(Y)$:

$$K(\phi_S(X), \phi_S(Y)) = \exp\left(-\frac{1}{2\gamma^2}\|(\phi_S(X) - \phi_S(Y))\|^2\right)$$

where $\|\ \|$ denotes Euclidean norm and $\gamma$ is the length-scale parameter. Since kernel measures the similarity between the two time-series we used the following formula to define our DTW kernel-based distance:

$$d_K(X, Y) = 1 - K(\phi_S(X), \phi_S(Y))$$

Now, we can replace DTW in the optimization problem (5) with the new distance, leading to the new formulation:

$$(a^*, b^*) = \underset{(a,b):1\leq a^*\leq b^*\leq M}{\operatorname{argmin}} d_K(X, Y_{a:b}) \tag{7}$$

Unfortunately, due to the nature of $d_K(X, Y)$ – and more precisely the computation of $\boldsymbol{\phi}_S(Y_{a:b})$ for every pair $(a, b)$ – the objective function of (7) is black-box and expensive. We have decided to use Bayesian optimization to solve (7), a sample efficient technique for optimizing black-box, multi-extremal and expensive objective functions (Frazier, 2018). It is successfully applied in the Machine Learning community for automating the configuration of Machine Learning algorithms – autoML (Hutter et al., 2019; Feurer et al., 2019) – as well as complex Machine Learning pipelines (Candelieri and Archetti, 2019).

Finally, to avoid undesired solutions, we have introduced "reasonable" constrains with respect to the length of subsequences found in $Y$:

$$\begin{bmatrix} 1 & -1 \\ -1 & 1 \end{bmatrix} \begin{bmatrix} a^* \\ b^* \end{bmatrix} \ll \begin{bmatrix} (1-v)N \\ (1+v)N \end{bmatrix}, \tag{8}$$

where $v$ is the admitted deviation, in percentage, between the length of the reference pattern and the length of an identified subsequence.

### 3.3. Kernel learning for user identification from walking activity

In this section we present how we have extended our approach to implement a specific application: the identification of a user from his/her walking activity data.

In this specific setting we assume to have $n$ different reference patterns $X_i$ ; $i = 1, \ldots, n$, one for each possible user to identify. The first step is to generate a specific kernel for each reference pattern – i.e., for each user – by learning the best value of the length-scale $\gamma_i$ directly from data:

$$\gamma^* = \underset{\gamma \in \mathbb{R}^n}{\operatorname{argmin}} \frac{2}{n(n-1)} \sum_{i=1}^{n-1} \sum_{j=i+1}^{n} \left| \left(1 - K_{\gamma_i}(\phi_S(X_i), \phi_S(X_j))\right) - DTW(X_i, X_j) \right| \tag{9}$$

where $\gamma = (\gamma_1, \ldots, \gamma_n)^{\mathrm{T}}$.

The problem (9) consists in minimizing the upper-triangular matrix of error between the exact DTW and the approximated distance $1 - K_{\gamma_i}(\phi_S(X_i), \phi_S(X_j))$. Since the range of RBF kernel is in [0,1], DTW is preliminary rescaled in the same interval.

In the following we summarize the procedure to construct the learning kernel-based algorithm enable user identification through rough data streams given by accelerometers. For simplicity we assumed equal number of the patterns and data streams, in general it can be suited to any size and number of time series to compare.

---

**User identification from walking activity algorithm – trough kernel-based DTW approximation**

**Input**: $n$ reference patterns $\{X_i\}_{i=1\ldots n}$, $m$ data streams $\{Y_i\}_{i=1\ldots m}$ and a hyperparameters $R, D_{min}, D_{max}, v, \sigma^2$.
**Output**: a matrix $\mathbf{K} \in \mathbb{R}^{n \times n}$ containing the kernel values.
1:    generate a set of $R$ "basis" time series $S = \{s_1, s_2, \ldots, s_R\}$, where each $s_i \sim \mathcal{N}(0, \sigma^2)$
2:    **for** i=1:$n$ **do**
3:       compute $\phi_S(X_i) = \left(DTW(X_i, s_1), \ldots, DTW(X_i, s_R)\right)^T$
4:    **endfor**
5:    $\mathbf{M}_{i,j} = DTW(X_i, X_j), \forall i, j = 1, \ldots, n$
6:    normalize entries of the matrix **M**
7:    choose $\gamma^* = \operatorname{argmin}_{\gamma \in \mathbb{R}^n} \frac{2}{n(n-1)} \sum_{i=1}^{n-1} \sum_{j=i+1}^{n} \left| \left(1 - K_{\gamma_i}(\phi_S(X_i), \phi_S(X_j))\right) - DTW(X_i, X_j) \right|$
8:    compute every entry of **K** as $\mathbf{K}_{i,j} = d_{K,\gamma^*}(X_i, Y_{a^*:b^*}^j)$, where $a^*, b^*$ are obtained solving (7) subject to constrains in (8)

---

## 4. Experimental setting

### 4.1. A benchmark dataset

The dataset we considered for the experiment is a well-known benchmark dataset for "user identification from walking activity" (Casale et al., 2012), which can be freely downloaded from the UCI Repository website (https://archive.ics.uci.edu/ml/datasets/User+Identification+From+Walking+Activity) .

The dataset refers to accelerometer data (i.e., acceleration on the x, y and z axes) acquired through an Android smartphone positioned in the chest pocket and from 22 participants walking in the wild over a predefined path.

Data information:

- Sampling frequency of the accelerometer: DELAY_FASTEST with network connections disabled
- A separate file for each participant
- Each file contains the following information
- Every row in each file consists of time-step, x acceleration, y acceleration, z acceleration

The dataset was intended for identification and authentication of people using motion patterns; in this study we adopted it to validate and compare a DTW- and a kernel-based solution.

We have selected, for every user, a recording window of 200 samples (around 6 seconds) and used this window as reference pattern (i.e., user's walking activity "signature"). The widows have been selected after some preliminary exploratory analysis on the entire set of recordings.

## 5. Results

In the following we report the main results of the study and the comparison between the user identification provided by our kernel-based approach with respect to the traditional DTW. As far as our approach is concerned, we have set the following values for our algorithm's parameters: $R = 64$, $D_{min} = 20$, $D_{max} = 30$, $\nu = 0.5$ and $\sigma^2 = 0.4$.

The following picture summarizes the level of association between each reference pattern – rows of the matrix – and subsequence identified within each data stream – columns of the matrix. The brighter the colour in a cell the lower the kernel-based distance. Therefore, bright colours on the diagonal represents correct identifications (17 out of 22, around 77%) while bright cell outside the diagonal are identification errors.

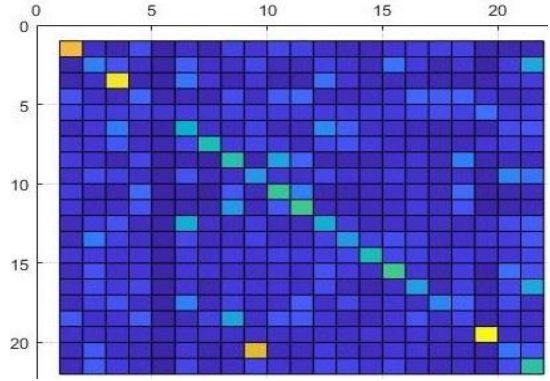

**Fig. 3.** Kernel-based distance between every reference pattern and the associated subsequences identified within the data streams: row $i$ refers to the $i$-th data stream and column $j$ refers to the $j$-th reference pattern. The brighter the colour the lower the distance.

The following figure summarizes the same kind of results but obtained through a traditional DTW analysis, in this case the correct identifications are 19 out of 22 data streams (around 86%).

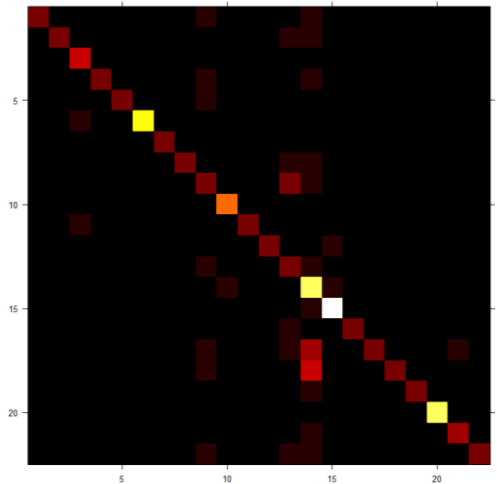

**Fig. 4.** DTW between every reference pattern and the associated subsequences identified within the data streams: row $i$ refers to the $i$-th data stream and column $j$ refers to the $j$-th reference pattern. The brighter the colour the lower the DTW value.

In both cases identification errors occur: a common source of error could be our arbitrary selection of the reference patterns.

Our kernel-based approach has a higher error due to its approximating nature; however, this higher error is counterbalanced by the significant reduction in computational complexity.

# 6. Conclusions

Computational complexity of DTW was addressed, in this paper, through an approach aimed at learning a kernel to approximate it. Contrary to DTW, kernel requires that the two elements – in this case time series – to be compared has the same length; a random feature embeddings have been used in order to map any time series in a space spanned by a given number of "basis" time series, randomly chosen. Thanks to this idea, the approach can be also applied to solve subsequence search, where one of the time series is a reference pattern and the other is a longer data streams where the reference pattern is searched for into.

Validation was performed on a benchmark dataset related to the user identification from walking activity data. The kernel-based DTW approximation proved to be effective and comparable, in terms of accuracy, to a traditional DTW solution. As expected, due to its approximating nature, the accuracy was slightly lower, but the improvement in terms of computational cost was more relevant. The gain depends on different factors, more precisely the number of "basis" time series, their (maximum) length and the possibility to parallelize the computation of the components of the features map. As ongoing work, we are extending the experiments to estimate the impact of these algorithm's parameters onto accuracy and computational time. Furthermore, on the benchmark dataset, we are planning to extend the subsequence search to the identification of multiple repetitions, counting the "compliant" repetitions and classify the data stream (i.e., associated the data stream to a specific user) depending on their number. The aim is to further increase accuracy, despite some increase in the computational cost.

Finally, we are also working on validating the approach on a real-life case study that have already addressed this topic via traditional DTW (Candelieri et al., 2018), consisting in the evaluation of physical rehabilitation in elderly people. The aim is to evaluate the capabilities of the kernel-based approximation DTW in a real-life setting.

**Funding:** This research leading to these results has received funding from The Home of Internet of Things (Home IoT), CUP: E47H16001380009 - Call "Linea R&S per Aggregazioni" cofunded by POR FESR 2014-2020.